%% file: root.tex
\documentclass[letterpaper, 10 pt, conference]{ieeeconf}  

\IEEEoverridecommandlockouts                              

\title{\LARGE \bf
Situational Confidence Assistance for Lifelong Shared Autonomy
}

\author{Matthew Zurek$^*$, Andreea Bobu$^*$, Daniel S. Brown, and Anca D. Dragan
\thanks{*Indicates equal contribution.
Authors are with EECS at UC Berkeley. 
Research supported by the Air Force Office of Scientific Research (AFOSR), the Office of Naval Research (ONR), and
NSF grant IIS1734633 (SCHooL).}
}

\usepackage{xcolor}
\usepackage{cite}
\usepackage{amsmath} 
\usepackage{amssymb}  
\usepackage{stackengine}
\stackMath
\usepackage{cite}
\usepackage[ruled,noend]{algorithm2e}
\usepackage{graphicx}
\usepackage{caption}
\usepackage{subcaption}
\usepackage[switch]{lineno}

\input{macros}
\usepackage[normalem]{ulem}
\newcommand{\adnote}[1]{\textcolor{red}{AD: #1}}

\newcommand{\abnote}[1]{\textcolor{blue}{AB: #1}}

\newcommand{\comment}[1]{}

\usepackage{xspace}
\newcommand{\algname}{Confidence-Aware Shared Autonomy\xspace}
\newcommand{\algabbr}{CASA\xspace}

\begin{document}

\maketitle
\thispagestyle{empty}
\pagestyle{empty}

\comment{

\begin{abstract}
Robots can assist people to accomplish tasks better than they can perform them on their own.
However, a robot can only be helpful if it can infer the human's intent.
Existing shared autonomy algorithms rely on a predefined set of skills and seek to assist the human towards the most likely one from this set. This is a problem if the desired skill is not known by the robot because then the assistance algorithm will hinder the human's performance by assisting for the wrong thing.
We present a general method for continual learning and assistance via shared autonomy. Our method provides the robot with situational confidence that allows it to realize when it does not have a good explanation for the human's behavior. Rather than assisting for the wrong skill, our method allows the robot to control its assistance in a way that is proportional to its confidence that the human's intent corresponds to a known skill.
We demonstrate with both an expert case study and a user study that our proposed method achieves better task performance with less effort than prior approaches for shared autonomy. Our method quickly recognizes when the human is attempting a novel task, learns the new skill from human demonstrations, and then adds this new skill to its library of known skills, enabling efficient lifelong learning for confidence-based shared autonomy.
\end{abstract}
}

\begin{abstract}

Shared autonomy enables robots to infer user intent and assist in accomplishing it. But when the user wants to do a new task that the robot does not know about, shared autonomy will hinder their performance by attempting to assist them with something that is not their intent. Our key idea is that the robot can detect when its repertoire of intents is insufficient to explain the user's input, and give them back control. This then enables the robot to observe unhindered task execution, learn the new intent behind it, and add it to this repertoire. We demonstrate with both a case study and a user study that our proposed method maintains good performance when the human's intent is in the robot's repertoire, outperforms prior shared autonomy approaches when it isn't, and successfully learns new skills, enabling efficient lifelong learning for confidence-based shared autonomy.

\end{abstract}

\input{1_intro}
\input{2_method}

\input{3_casestudy}
\input{4_expertstudy}
\input{5_discussion}

{
\bibliographystyle{IEEEtran}
\bibliography{IEEEabrv,references}}

\end{document}

%% file: macros.tex
\newcommand{\sysstate}{x}
\newcommand{\action}{u}
\newcommand{\hinput}{a}
\newcommand{\goal}{\ensuremath{\theta}}

\newcommand{\aset}{\ensuremath{\mathcal{U}}}
\newcommand{\sset}{\ensuremath{\mathcal{X}}}
\newcommand{\gset}{\ensuremath{\Theta}}
\newcommand{\hset}{\ensuremath{\mathcal{A}}}

\newcommand{\cost}{C}
\newcommand{\traj}{\ensuremath{\xi}}

%% file: 1_intro.tex
\section{Introduction}
\label{sec:intro}


In shared autonomy \cite{aigner1997human,dragan2013blending,reddy2018shared,losey2018review,goertz1963manipulators,farraj2018hapticshared,losey2020latent,li2011dynamic,erdogan2017effect,brown2014balancing,crandall2017human}, robots assist human operators to perform their objectives more effectively. Here, rather than directly executing the human's control input, a typical framework has the robot estimate the human's intent and execute controls that help achieve it \cite{dragan2013blending,javdani2015shared,muelling2017autonomy,perez2015fast,reddy2018shared}.


These methods succeed when the robot knows the set of possible human intents a priori, e.g. the objects the human might want to reach, or the buttons they might want to push \cite{dragan2013blending,javdani2015shared}. But realistically, users of these systems will inevitably want to perform tasks outside the repertoire of known intents -- they might want to reach for a goal unknown to the robot, or perform a new task like pouring a cup of water into a sink. This presents a three-fold challenge for shared autonomy. First, the robot will be unable to recognize and help with something unknown. Second, and perhaps more importantly, it will attempt to assist with whatever wrong intent it infers, interfering with what the user is trying to do and hindering their performance. This happens when the robot plans in expectation \cite{javdani2015shared}, and, as our experiments will demonstrate, it happens even when the robot arbitrates the amount of assistance based on its confidence in the most likely goal \cite{dragan2013blending}. Third, the new task remains just as difficult as the first time even after arbitrarily many attempts.

Our key idea is that the robot should detect that the user is trying something new and give them control. This then presents an opportunity for the robot to observe the new executed trajectory, learn the underlying intent that explains it, and add it to its repertoire so that it can infer and assist for this intent in the future.

\begin{figure}[t]
\includegraphics[width=\linewidth]{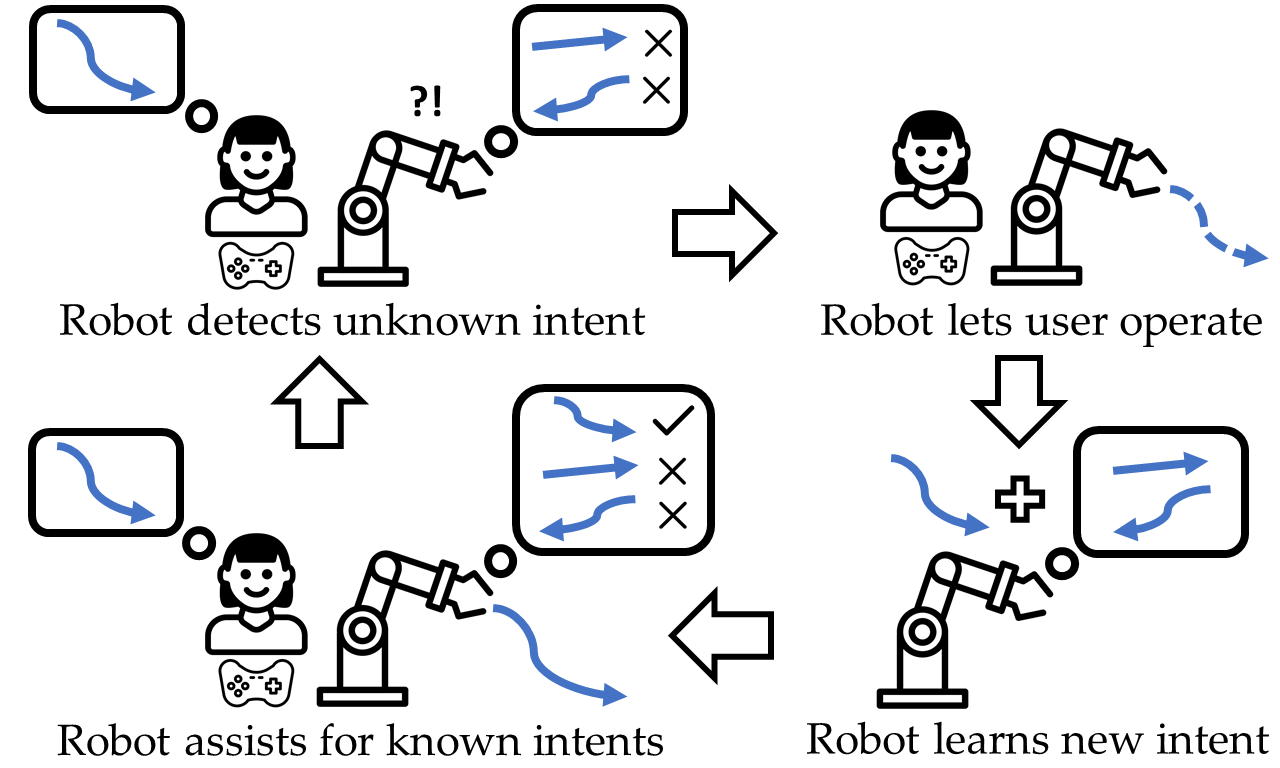}
\centering
\caption{We propose an approach for lifelong shared autonomy that enables a robot to detect when its set of known human intents is insufficient to explain the current human behavior. Rather than trying to assist for the wrong intent, the robot learns from novel teleoperations to learn a model of the new intent, allowing for lifelong confidence-based assistance.}
\label{fig:front_fig}
\vspace*{-0.5cm}
\end{figure}

To achieve this, we need two ingredients: 1) a way for the robot to detect its repertoire of intents is insufficient, and 2) a representation of intents that enables learning new tasks throughout its lifetime, adding them to its repertoire, and performing inference over them in a unified way with the initial known intents. 
For the latter, we use cost functions to unify goals and general skills like pouring into the same representation.
This then enables the former: when the human acts too suboptimally for any of the known cost functions, it suggests the robot lacks the correct set of costs.

Our approach takes inspiration from recent work on hypothesis misspecification where the robot recognizes when its cost function features are insufficient to explain human demonstrations and corrections \cite{bobu2020quantifying}, and updates the cost in proportion to the \textit{situational confidence} in these features' ability to explain input. We extend detecting hypothesis mispecification to the context of shared autonomy, in which there are multiple intents, represented as cost functions, and the robot seeks to recognize whether any of the known intents explain the human input sufficiently. The robot can then arbitrate its assistance based on its confidence in the most likely intent being what the human wanted.

Our approach, which we call Confidence-Aware Shared Autonomy (CASA), allows the robot to ascertain whether the human inputs are associated with a known or new task. 
By arbitrating the user's input based on the confidence in the most likely intent, \algabbr follows a standard policy blending assistance approach if the task is known, and otherwise gives the user full control. Additionally, \algabbr allows the user to provide a few demonstrations of the new intent, which the robot uses to learn a cost function via Inverse Reinforcement Learning (IRL) \cite{finn2016guided} and add it to its set of intents. This enables lifelong shared autonomy, where the robot helps when it is confident in what the user wants and learns new intents when it detects that the human is doing something novel, so that it can assist with that intent in the future.

We test our approach in a expert case study and a user study with a simulated 7DoF JACO assistive robot arm.
Our results suggest that \algabbr significantly outperforms prior approaches when assisting for unknown intents, maintains high performance in the case of known ones, and successfully learns new intents for better lifelong shared autonomy.

%% file: 2_method.tex
\section{Confidence-Aware Shared Autonomy}
\label{sec:method}

We consider a human teleoperating a dexterous robotic manipulator to perform everyday manipulation tasks.
The robot's goal is to assist the person in accomplishing their desired skill by augmenting or changing their input. 
While the robot possesses a predefined set of possible intents, the human's desired motion might not be captured by any of them.
We propose that since the robot might not understand the person's intentions, it should reason about how confident it is in its predictions to avoid assisting for the wrong intent.

\subsection{Preliminaries}

Formally, let $\sysstate\in\sset$ be the continuous robot state (e.g. joint angles), and $\action\in\aset$ the continuous robot action (e.g. joint velocity).
The user controls their desired robot configuration by providing continuous inputs $\hinput\in\hset$ via an interface (e.g. GUI, joystick, keyboard commands, etc). These inputs are mapped to robot actions through a \textit{direct teleoperation} function $\mathcal{T} : \hset \rightarrow \aset$. Define a person's trajectory up until time $t$ as the sequence $\traj_{0\to t} = (\sysstate^0, \hinput^0, \hdots, \sysstate^t, \hinput^t)$.

The robot is equipped with a set of known intents $\gset$, one of which may represent the user's desired motion. Each intent is parameterized by a cost function $\cost_\goal$, which may be hand-engineered or learned from demonstrations via IRL~\cite{maxent,Ng2000inverse}. For example, if the intent represents moving to a goal $g$, the cost function can be distance to the goal: $\cost_g(\traj)=\sum_{x\in\traj}\|x-g\|$. If the intent is pouring a cup, the cost can be a neural network with parameters $\psi$, $\cost_\psi$. 
Our shared autonomy system does not know the intent a priori, but infers it from the human's inputs. 
Given the user's trajectory so far, $\traj_{0\to t}$, a common strategy is to predict the user's intent $\goal\in\gset$, compute the optimal action for moving accordingly, then augment the user's original input with it~\cite{dragan2013blending}.

However, what if none of the intents match the human's input, i.e., the person is trying to do something the robot does not know about?
We introduce a shared autonomy formalism where the robot reasons about its confidence in its current set of intents' ability to explain the person's input, and uses that confidence for robust assistance. This confidence serves a dual purpose, as the robot can also use it to ask the human to demonstrate 
the missing intent.

\subsection{Intent Inference}
\label{sec:skill_inference}

To assist the person, the robot has to first predict which of its known tasks the person is trying to carry out, if any.
To do that, the robot needs a model of how people teleoperate it to achieve a desired motion. 
We assume the Boltzmann noisily-rational decision model~\cite{baker2007goal,von1945theory}:
\begin{equation}
    P(\traj \mid \goal, \beta) = \frac{e^{-\beta \cost_{\goal}(\traj)}}{\int_{\bar{\traj}}e^{-\beta \cost_{\goal}(\bar{\traj})} d\bar{\traj}} \enspace ,
    \label{eq:boltzman_trajs}
\end{equation}
where the person chooses the trajectory $\traj$ proportional to its exponentiated cost $\cost_\goal$. The parameter $\beta \in [0, \infty)$ controls how much the robot expects to observe human input consistent with the intent $\goal$. Typically, $\beta$ is fixed, recovering the Maximum Entropy IRL observation model~\cite{maxent}, which is what most inference-based shared autonomy methods use~\cite{dragan2013blending,javdani2015shared}.
Inspired by work on confidence-aware human-robot interaction~\cite{fridovich-keil2019confidence,fisac2018probabilistically,bobu2020quantifying}, we instead reinterpret $\beta$ as a measure of the robot's \textit{situational confidence} in its ability to explain human data, given the known intents $\gset$,
and we show how the robot can estimate it in Sec. \ref{sec:confidence_estimation}. 

Given Eq.~\eqref{eq:boltzman_trajs}, if the cost $\cost_\goal$ of intent $\goal$ is additive along the trajectory $\traj$, we have that:
\begin{equation}
P(\traj_{0 \to t}\mid \goal, \beta) = e^{-\beta \cost_\goal(\traj_{0 \to t})} \frac{\int_{\bar\traj_{t \to T}} e^{-\beta \cost_\goal(\bar\traj_{t \to T})} }{\int_{\bar\traj_{0 \to T}} e^{-\beta \cost_\goal(\bar\traj_{0 \to T})} }\enspace,
\end{equation} 
where $T$ is the duration of the episode. In high-dimensional manipulation spaces, evaluating these integrals is intractable. We follow \cite{dragan2013blending} and approximate them via Laplace's method:
\begin{eqnarray}
P(\traj_{0 \to t}\mid \goal, \beta) \approx e^{-\beta \left(\cost_\goal(\traj_{0 \to t}) + \cost_\goal(\traj^*_{t \to T}) - \cost_\goal(\traj^*_{0 \to T})\right)} \nonumber \\
\times \sqrt{\left(\frac{\beta}{2\pi}\right)^{tk}\frac{|\nabla^2 \cost_\goal(\traj^*_{0 \to T})|}{|\nabla^2 \cost_\goal(\traj^*_{t \to T})|}}\enspace,
\label{eq:laplace}
\end{eqnarray}
where $k$ is the action dimensionality, and the trajectories $\traj^*_{0 \to T}$ and $\traj^*_{t \to T}$ are optimal with respect to $\cost_\goal$ and can be computed with any off-the-shelf trajectory optimizer\footnote{We use TrajOpt~\cite{trajopt}, based on sequential quadratic programming.}.

Now, given a tractable way to compute the likelihood of the human input, the robot can obtain a posterior over intents:
\begin{equation}
    P(\goal \mid \traj_{0 \to t}, \beta) = \frac{P(\traj_{0 \to t}\mid \goal, \beta)}{\sum_{\goal' \in \gset} P(\traj_{0 \to t}|\goal', \beta)},
    \label{eq:goal_inference}
\end{equation}
assuming $P(\goal\mid\beta)=P(\goal)$ and a uniform prior over intents. 

Prior inference-based shared autonomy work~\cite{dragan2013blending,javdani2015shared} typically assumes $\beta=1$.
We show that the robot should not be restricted by such an assumption and it, in fact, benefits from estimating $\hat\beta$ and reinterpreting it as a confidence.

\subsection{Confidence Estimation}
\label{sec:confidence_estimation}

In the Boltzmann model in Eq.~\eqref{eq:boltzman_trajs}, we see that $\beta$ determines the variance of the distribution over human trajectories. When $\beta$ is high, the distribution is peaked around those trajectories $\xi$ with the lowest cost $\cost_\goal$; in contrast, a low $\beta$ makes all trajectories equally likely. We can, thus, reinterpret $\beta$ to take a useful meaning in shared autonomy: given an intent, $\beta$ controls how well that intent's cost explains the user's input. A high $\beta$ for an intent $\goal$ indicates that the intent's cost explains the input well and is a good candidate for assistance. A low $\beta$ on all intents suggests that the robot's intent set is insufficient for explaining the person's trajectory.

We can thus estimate $\beta$ and use it for assistance. Using the likelihood function in Eq.~\eqref{eq:laplace}, we write the $\beta$ posterior
\begin{equation}
    P(\beta \mid \traj_{0 \to t}, \goal) = \frac{P(\traj_{0 \to t}\mid \goal, \beta)P(\beta)}{\int_{\bar\beta} P(\traj_{0 \to t}|\goal, \bar\beta)P(\beta)d\bar\beta}.
\end{equation}
%
If we assume a uniform prior $P(\beta)$, we may compute an estimate of the confidence parameter $\beta$ per intent $\goal$ via a maximum likelihood estimate:
\begin{equation}\label{eq:max_lik_beta}
     \hat\beta_\goal=\arg\max_{\bar\beta} e^{-\bar\beta \left(\cost_\goal(\traj_{0 \to t}) + \cost_\goal(\traj^*_{t \to T}) - \cost_\goal(\traj^*_{0 \to T})\right)}
 \left(\frac{\bar\beta}{2\pi}\right)^{\frac{tk}{2}}\enspace,
\end{equation}
%
where we drop the Hessians since they don't depend on $\beta$. Setting the derivative of the objective in Eq.~\eqref{eq:max_lik_beta} to zero and solving for $\beta$ yields the following estimate:
\begin{equation}\label{eq:beta_mle}
\hat\beta_{\goal}^{MLE} = \frac{tk}{2(\cost_\goal(\traj_{0 \to t}) + \cost_\goal(\traj^*_{t \to T}) - \cost_\goal(\traj^*_{0 \to T})) }\enspace.
\end{equation}
Alternatively, we chose to add an exponential prior with parameter $\lambda$, $Exp(\lambda)$, on $\beta$ to obtain a MAP estimate 
\begin{equation}\label{eq:beta_map}
    \hat\beta_{\goal}^{MAP} = \frac{tk}{2( \lambda + \cost_\goal(\traj_{0 \to t}) + \cost_\goal(\traj^*_{t \to T}) - \cost_\goal(\traj^*_{0 \to T})) }\enspace.
\end{equation}
%
The denominators in equations \ref{eq:beta_mle} and \ref{eq:beta_map} can be interpreted as the ``suboptimality'' of the observed partial trajectory $\traj_{0 \to t}$ compared to the cost of the optimal trajectory for the particular $\goal$, $ \cost_\goal(\traj^*_{0 \to T})$. Note that
$\hat\beta_\goal$ is inversely proportional to the suboptimality divided by the number of time steps $t$ that have passed. 
Intuitively, the user has more chances to be a suboptimal teleoperator as time goes on, so dividing for $t$ corrects for the natural increase in suboptimality over time. 

If this normalized suboptimality is low for an intent $\goal$, then the person is close to a good trajectory for that intent and $\hat\beta_\goal$ will be high. Thus, a high $\hat\beta_\goal$ means that the person's input is well-explained by that intent.
On the other hand, high suboptimality per time means the person is far from good trajectories, so $\goal$'s cost model $\cost_\goal$ does not explain the person's trajectory and $\hat\beta_\goal$ will be low.



\comment{
\begin{figure*}
\centering
\begin{subfigure}{.25\textwidth}
  \centering
  \includegraphics[width=\textwidth,clip=false]{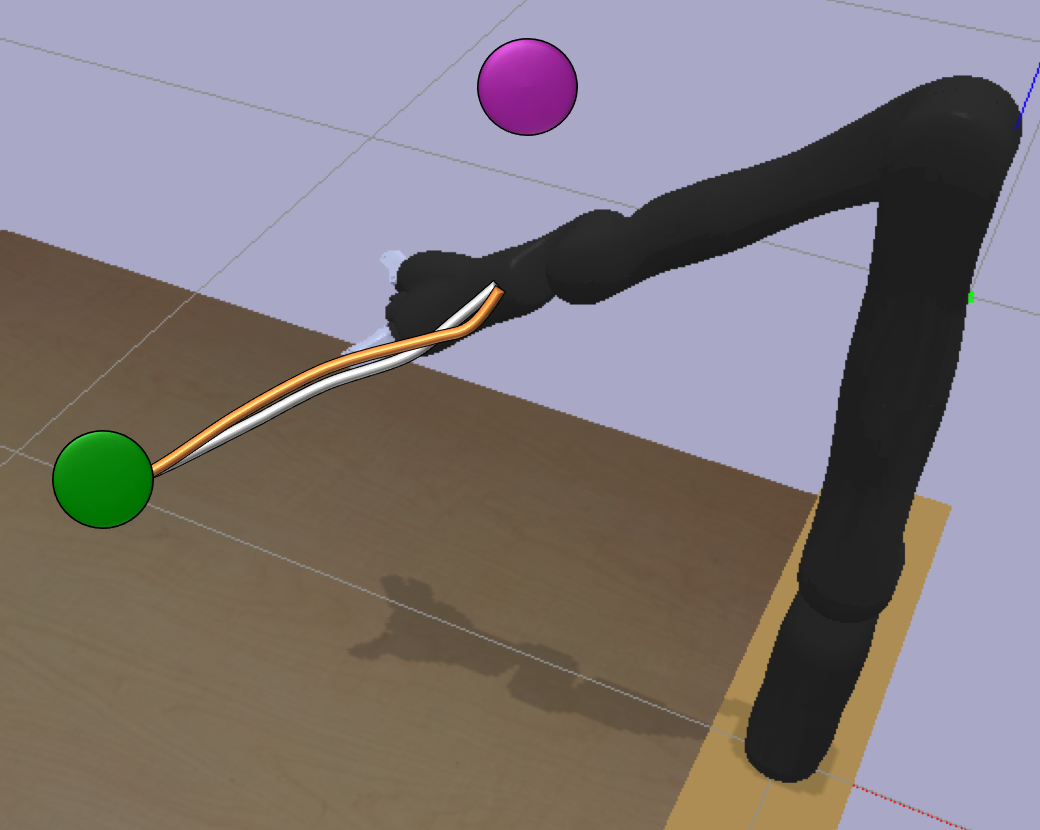}
\end{subfigure}
\centering
\begin{subfigure}{.7\textwidth}
  \centering
  \includegraphics[width=\textwidth,clip=false]{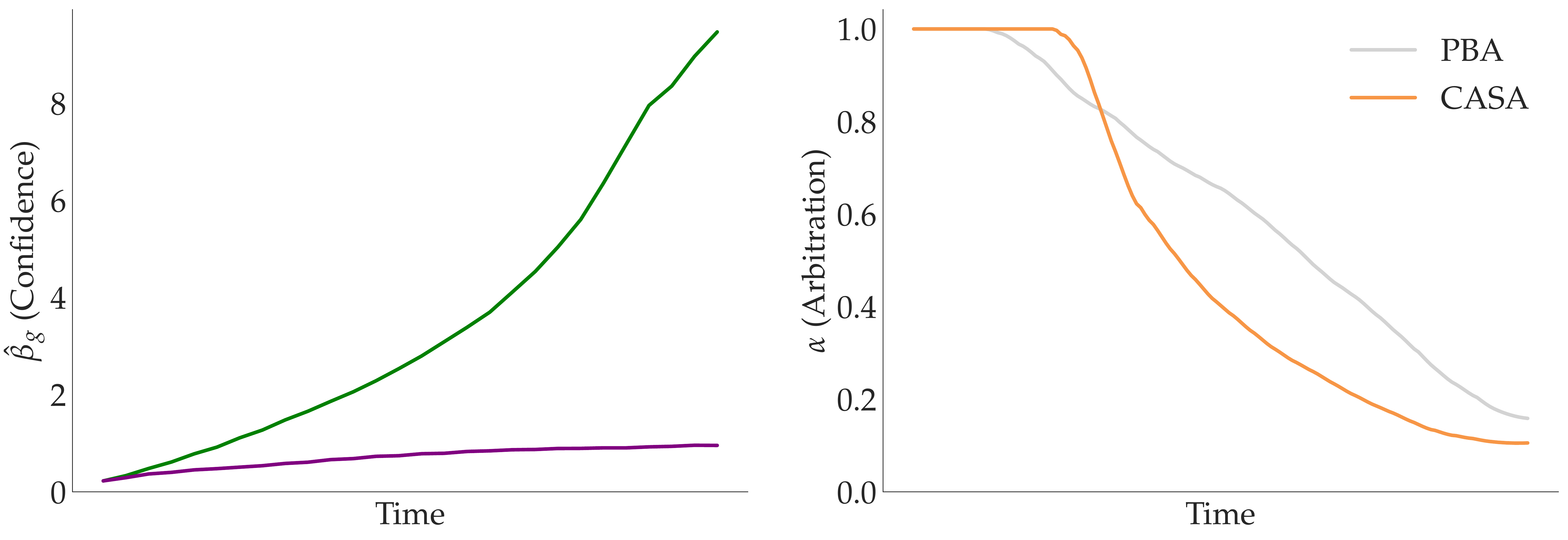}
\end{subfigure}
\centering
\begin{subfigure}{.25\textwidth}
  \centering
  \includegraphics[width=\textwidth,clip=false]{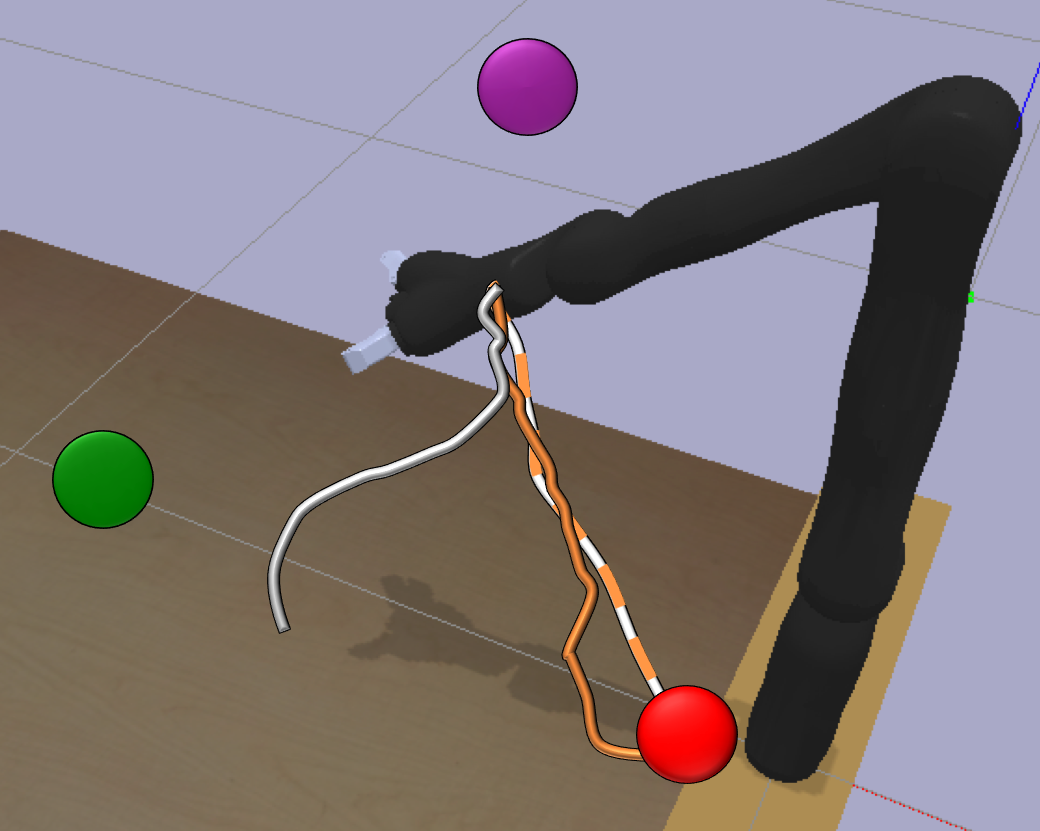}
\end{subfigure}
\centering
\begin{subfigure}{.7\textwidth}
  \centering
  \includegraphics[width=\textwidth,clip=false]{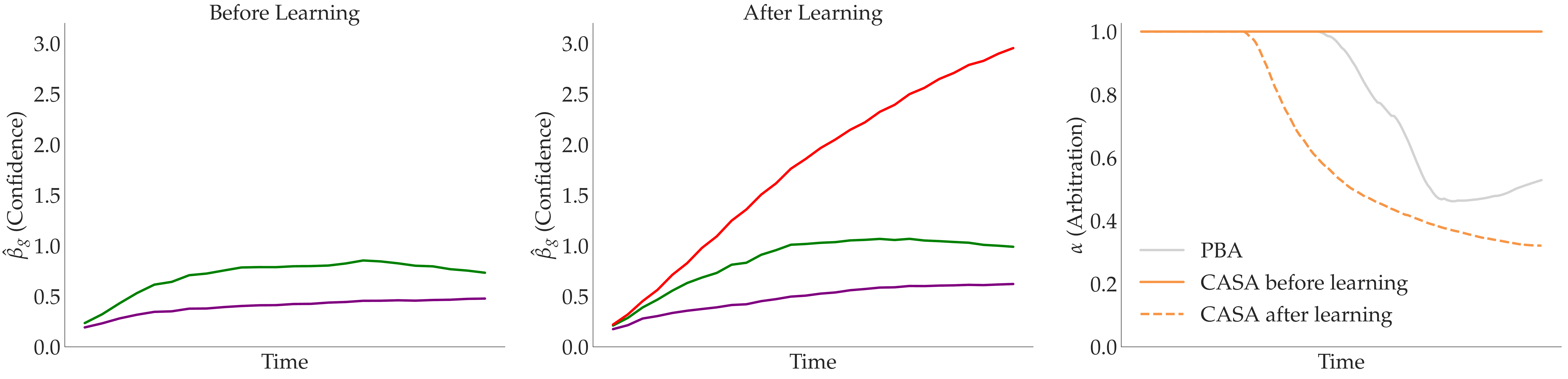}
\end{subfigure}
\caption{TODO.\abnote{update firgure to include pouring,write caption, update text referring to the figure}\adnote{add labels maybe on the left hand side in vertical text to say know goal, unknown goal, unknown skill (pouring)}}
\label{fig:case_study}
\end{figure*}
}

\begin{figure*}
\centering
\includegraphics[width=0.9\textwidth]{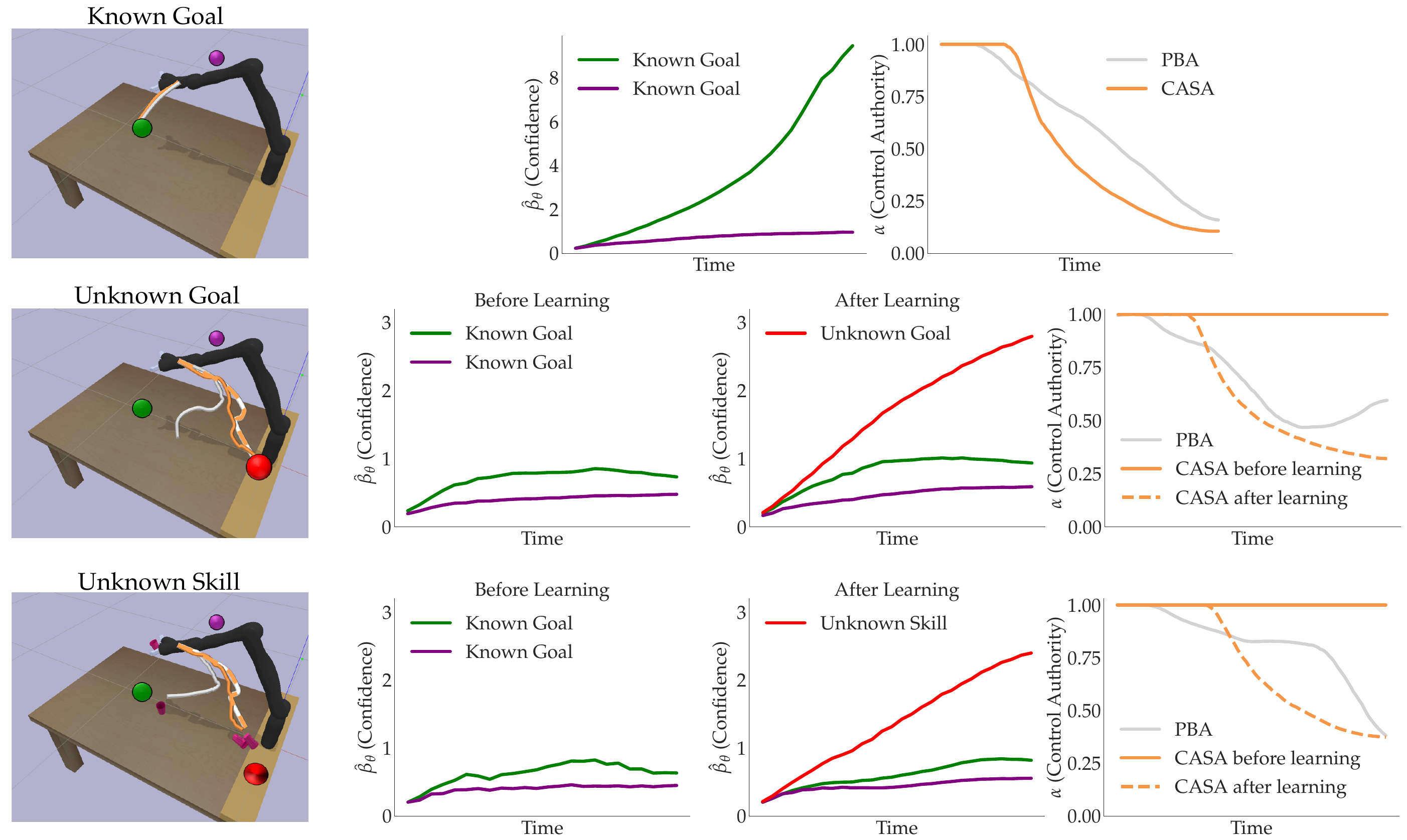}
\caption{Expert case study results. For each of three manipulation tasks, we compute confidence estimates before learning and, for the misspecified tasks (middle, bottom), we recompute the confidence estimates after learning. We also plot the strength of assistance before and after learning and compare to a policy blending baseline~\cite{dragan2013blending}.}
\label{fig:case_study}
\vspace*{-0.5cm}
\end{figure*}

\subsection{Confidence-based Arbitration}

\comment{
Armed with an estimate of the confidence $\hat\beta_\goal$ for every intent $\goal\in\gset$, the robot can now predict the most likely intent $\goal^* = \arg\max_{\goal\in\gset} P(\goal\mid \traj_{0\to t},\hat\beta_\goal)$ using Eq. \eqref{eq:goal_inference}. 
To assist for it, the robot reasons about how the person would like the task to be executed by computing an optimal trajectory $\traj^*=\arg\min_\traj\sum_{\sysstate\in\traj}\cost_\goal^*(\sysstate)$, and chooses the first action $\action^*$.

Now, given the robot's predicted next action $\action^*$ and the human's original action $\hinput^t$, the shared autonomy system has to combine them to best assist the human.
One way to so is by using the ``policy blending'' style of assistance~\cite{dragan2013blending}, which combines $\action^*$ and $\mathcal{T}(\hinput^t)$ using a blending parameter $\alpha\in[0,1]$, resulting in the robot action $\action^t = \alpha \mathcal{T}(\hinput^t) + (1-\alpha) \action^*$.
}

Armed with a confidence estimate $\hat\beta_\goal$ for every $\goal\in\gset$, the robot can predict the most likely one $\goal^* = \arg\max_{\goal\in\gset} P(\goal\mid \traj_{0\to t},\hat\beta_\goal)$ using Eq. \eqref{eq:goal_inference}.
From here, one natural style of assistance is ``policy blending'' ~\cite{dragan2013blending}.
First the robot computes an optimal trajectory under the most likely intent, $\traj^*=\arg\min_\traj\sum_{\sysstate\in\traj}\cost_\goal^*(\sysstate)$, the first action of which is $\action^*$. 
Then the robot combines $\action^*$ and $\mathcal{T}(\hinput^t)$ using a blending parameter $\alpha\in[0,1]$, resulting in the robot action $\action^t = \alpha \mathcal{T}(\hinput^t) + (1-\alpha) \action^*$. We also refer to $\alpha$ as the human's control authority.

Prior work proposes different ways to arbitrate between the robot and human actions by choosing $\alpha$ proportional to the robot's distance to the goal or to the probability of the most likely goal \cite{dragan2013blending}. 
However, when using the probability $P(\goal^*\mid\traj)$, $\goal^*$ might look much better than the other intents, resulting in the robot wrongly assisting for $\goal^*$. Distance-based arbitration ignores the full history of the user's input and can only accommodate simple intents.

Instead, we propose that the robot should use its confidence in the most likely intent, $\hat\beta_{\goal^*}$, estimated according to Sec. \ref{sec:confidence_estimation}, to control the strength of its arbitration: 
%
\begin{equation}
    \action^t = \min(1, 1/\hat\beta_{\goal^*}) \mathcal{T}(\hinput^t) + (1-\min(1, 1/\hat\beta_{\goal^*})) \action^*
\end{equation}
When $\hat\beta_{\goal^*}$ is high, i.e. the robot is confident that the predicted intent $\goal^*$ can explain the person's input, $\alpha$ is low, giving the robot more influence through its action $\action^*$. When $\hat\beta_{\goal^*}$ is low, i.e. not even the most likely intent explains the person's input, $\alpha$ increases, giving the person's action $\hinput^t$ more authority.

\subsection{Using Confidence for Lifelong Learning}
\label{sec:IRL}

Estimating the confidence $\hat\beta_{\goal}$ also lets the robot detect \textit{misspecification} in $\gset$: if all estimated $\hat\beta_{\goal}$ for $\goal\in\gset$ are below a threshold $\epsilon$, the robot is missing the person's intent.

Once the robot has identified that its intent set is misspecified, it should ask the person to teach it. 
We represent the missing intent $\goal_\phi$ as a neural network cost parameterized by $\phi$ and learn it via deep maximum entropy IRL~\cite{finn2016guided}.
The gradient of the IRL objective with respect to the cost parameters $\phi$ can be estimated by: $\nabla_{\phi}\mathcal{L} \!\approx\! \frac{1}{|\mathcal{D}^*|}\sum_{\tau \in \mathcal{D}^*} \!\!\nabla_{\phi}\cost_{\phi}(\tau) \!-\! \frac{1}{|\mathcal{D}^{\phi}|}\sum_{\tau \in \mathcal{D}^{\phi}}\!\! \nabla_{\phi}\cost_{\phi}(\tau)$. $\mathcal{D^*}$ are (noisy) demonstrations of the person executing the desired missing intent via direct teleoperation, and
$\mathcal{D}^{\phi}$ are trajectories sampled from the $\cost_\phi$ induced near the optimal policy.

Once we have a new intent $\goal_\phi$, the robot updates its intent set $\gset \gets \gset \cup \goal_\phi$. The next time the person needs assistance, the robot can perform confidence estimation, goal inference, and arbitration as before, using the new library of intents. While the complexity scales linearly with $| \gset |$, planning can be parallelized across each intent.

Learned rewards fit naturally into our framework, allowing for a simple way to compare against the known intents. However, one could imagine adapting our method to the many other ways to learn an intent, from imitation learning~\cite{ho2016GAIL,reddy2020sqil}, to dynamic movement primitives~\cite{paraschos2013PMP}. For instance, if we parameterize intents via policies, we can derive a similar confidence metric based on probabilities of observed human actions under a stochastic policy, rather than costs.


%% file: 3_casestudy.tex
\section{Expert Case Study}
\label{sec:expert}

\begin{figure}
\includegraphics[width=\linewidth]{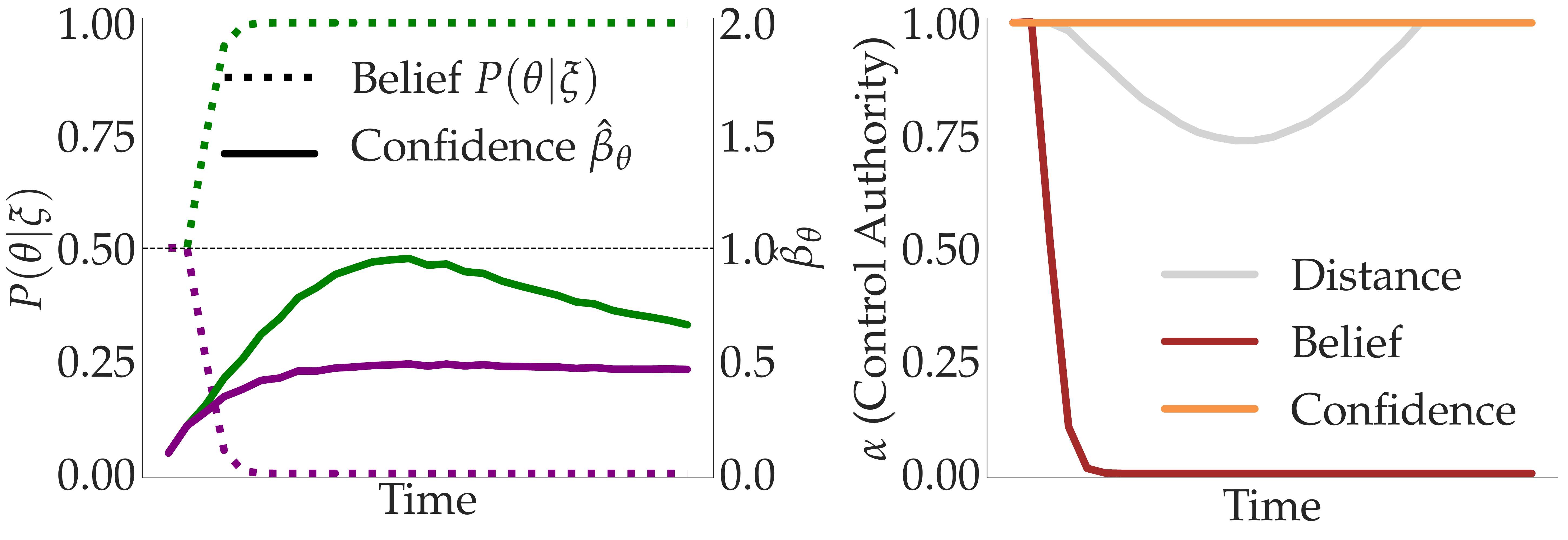}
\centering
\caption{Analysis of arbitration methods. After tracking an optimal trajectory for the Unknown Goal task, we show the robot's belief and confidence estimates for each known goal (left), as well as the $\alpha$ values under the distance, belief, and confidence-based arbitration schemes (right).}
\label{fig:alpha_methods}
\vspace*{-0.5cm}
\end{figure}

In this section, we introduce three manipulation tasks and use expert data to analyze confidence estimation and assistance.
We later put \algabbr's assistive capacity to test with non-experts in a user study in Sec. \ref{sec:userstudy}.

\subsection{Experimental Setting}

We conduct our experiments on the simulated 7-DoF JACO arm shown in Fig. \ref{fig:case_study}. We use the pybullet interface~\cite{coumans2019} and teleoperate the robot via keypresses. We map 6 keys to bi-directional $xyz$ movements of the robot's end-effector, and 2 keys for rotating it in both directions. We performed inference and confidence estimation twice per second.

We test \algabbr on 3 different tasks. 
In the \textit{Known Goal} task, we control for the well-specified setting: 
the robot must assist the user to move to the known green goal location in Fig. \ref{fig:case_study}. 
In the other tasks, we test CASA's efficacy in the case of misspecification, where the user's desired intent is initially missing from the robot's known set $\gset$.
In the second task, \textit{Unknown Goal}, the person teleoperates the robot to the red goal which is unknown to the robot. Finally, in the third and most complicated task, \textit{Unknown Skill}, the person tries to pour the cup contents at an unknown goal location. 

For the Unknown Goal and Unknown Skill tasks, we first run \algabbr before being exposed to the new intent (\algabbr \textit{before learning}). Detecting low confidence, the robot then asks for demonstrations and learns the missing intents via deep maximum entropy IRL as discussed in Sec. \ref{sec:IRL}. We then run teleoperation with \algabbr \textit{after learning}, to assess the quality of robot assistance after learning the new intent.

\begin{figure*}
\centering
\begin{subfigure}{.32\textwidth}
  \centering
  \includegraphics[width=\textwidth,clip=false]{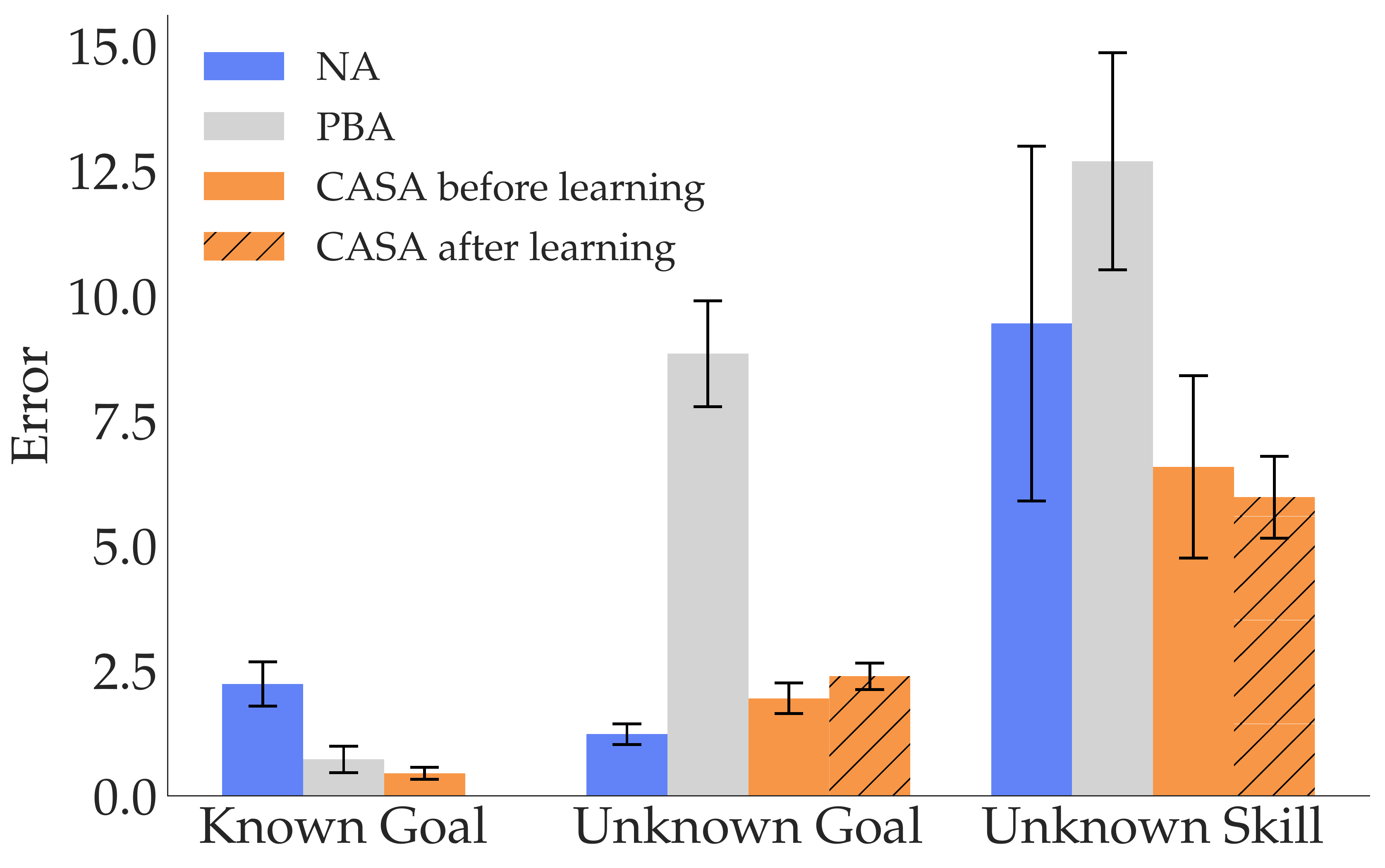}
\end{subfigure}
\centering
\begin{subfigure}{.32\textwidth}
  \centering
  \includegraphics[width=\textwidth,clip=false]{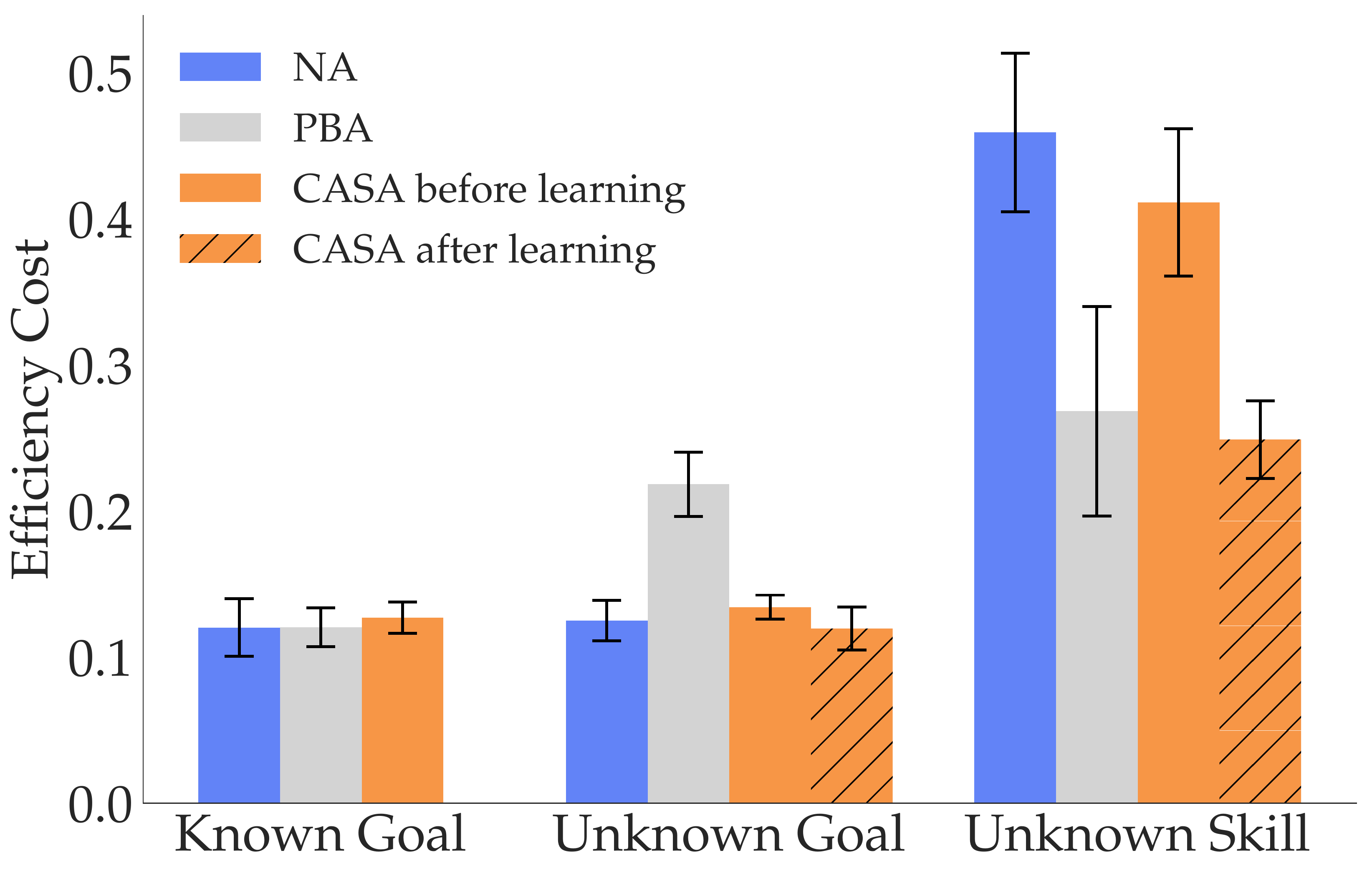}
\end{subfigure}
\centering
\begin{subfigure}{.32\textwidth}
  \centering
  \includegraphics[width=\textwidth]{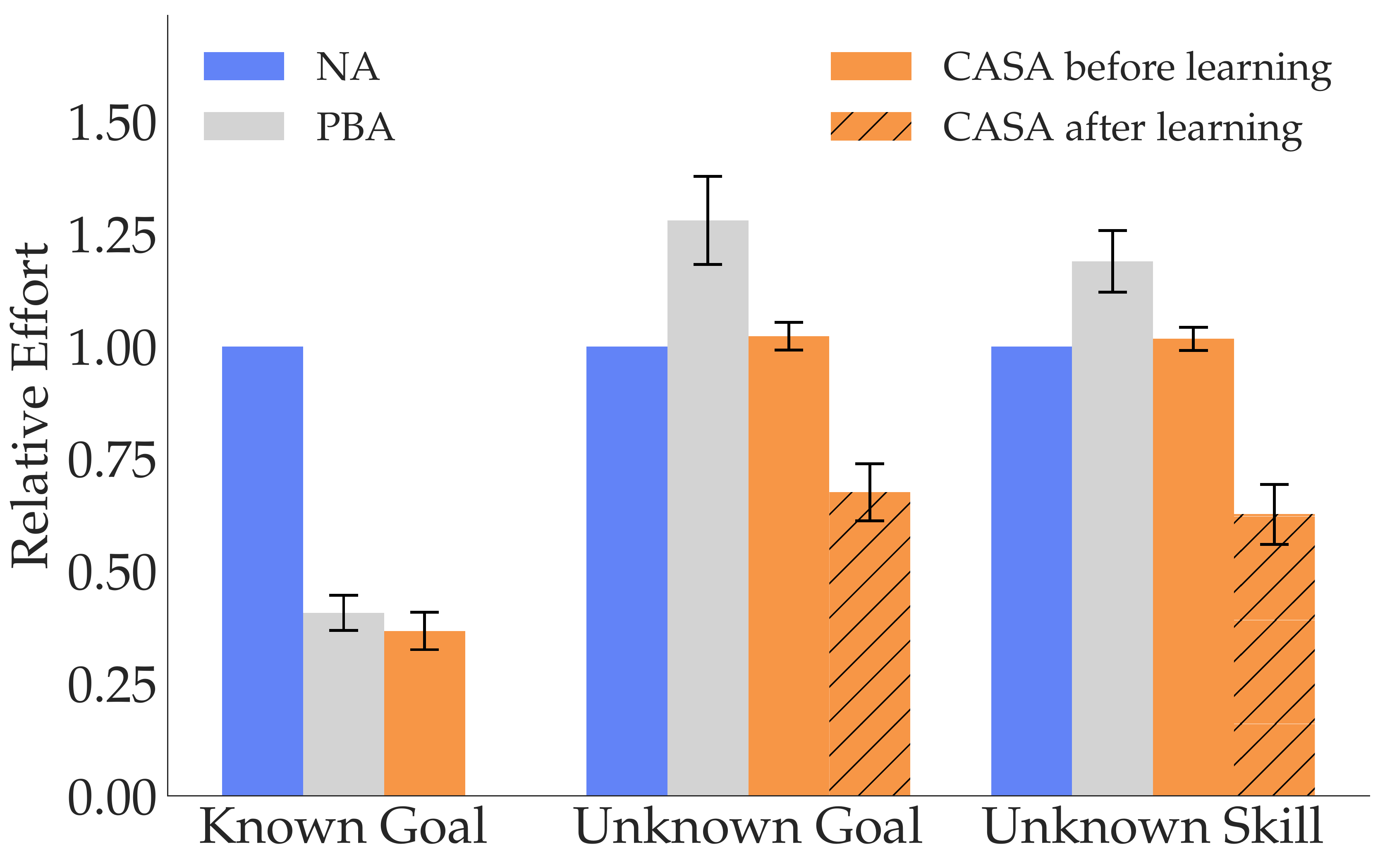}
\end{subfigure}
\caption{Our user study objective metrics. For every task, we measured error with respect to an intended trajectory (left), smoothness of the executed trajectory (middle), and effort relative to direct teleoperation (right).}
\label{fig:study_objective}
\vspace*{-0.5cm}
\end{figure*}

\subsection{Arbitration Method Comparison}
We compare \algabbr to a policy blending assistance (PBA) baseline \cite{dragan2013blending} that assumes $\beta=1$ for all intents. PBA arbitrates with the distance $d_{\goal^*}$ to the predicted goal: $\alpha = \min(1, d_{\goal^*}/D)$, with $D$ some threshold past which the robot does not assist. More sophisticated arbitration schemes use $P(\goal^* \mid \traj)$ or the full distribution $P(\goal \mid \traj)$, but they are much less robust to task misspecification. This is because when the user teleoperates for an unknown intent, $P(\traj \mid \goal)$ will be low for all known $\goal \in \gset$; however, forming $P(\goal \mid \traj)$ requires normalizing over all known intents, after which $P(\goal^* \mid \traj)$ can still be high unless the user happened to operate in a way that appears equally unlikely under all known intents.

We analyzed this phenomenon by tracking a reference trajectory for the Unknown Goal task which moves optimally towards the unknown goal (see Fig.~\ref{fig:case_study} for the task layout). We compared the performances of the distance and confidence arbitration methods, as well as a belief-based method which sets $\alpha = (P(\goal^* \mid \traj)|\gset| - 1)/(|\gset|-1)$ (chosen so that $\alpha= 0$ when $P(\goal^* \mid \traj) = 1/|\gset|$, $\alpha= 1$ when $P(\goal^* \mid \traj) = 1$). In Fig.~\ref{fig:alpha_methods}, the confidence in each goal stays low enough that the robot would have left the user in full control; meanwhile, the relatively higher likelihood of one goal causes the belief $P(\goal^* \mid \traj)$ to quickly go to $1$ and thus set the user's control authority to $0$ under the belief-based arbitration scheme.

We examined one belief-based arbitration method here, but since $P(\goal^* \mid \traj)$ rapidly goes to $1$, any other arbitration that is a function of the belief $P(\goal \mid \traj)$ would similarly try to assist for the wrong goal, motivating our choice of the simpler but more robust distance-based arbitration baseline. 

\subsection{Well-specified Tasks}

Fig. \ref{fig:case_study} (top) showcases the results of our experiment for the Known Goal task. Looking at the confidence plot, we see that $\hat\beta_\goal$ increases with time for the correct green goal, while it remains low for the alternate known purple goal. In the arbitration plot, as $\hat\beta_{\goal^*}$ increases, $\alpha$ gradually decreases, reflecting that the robot takes more control authority only as it becomes more confident that the person's intent is indeed $\goal^*$. Similarly, since there is no misspecification, PBA arbitration steadily decreases the human's contribution to the final control. Both methods result in smooth trajectories which go to the correct goal location. 

\subsection{Misspecified Tasks}

Our approach distinguishes itself in how it handles misspecified tasks. During the Unknown Goal task, in Fig. \ref{fig:case_study} (middle), \algabbr before learning estimates low $\hat\beta_\goal$ for both goals, since neither goal explains the person's motion moving towards the red goal. The estimated $\hat\beta_\goal$ is slightly higher for the green goal than for the purple one because it is closer to the user's input; however, neither are high enough to warrant an arbitration $\alpha$ below $1$, and thus the robot receives no control. In Fig. \ref{fig:case_study} (bottom), we observe almost identical behavior before learning for the Unknown Skill task: the known intents do not match the user's behavior, and thus the user is given full control authority and completes the task.

This contrasts PBA, which, for both Unknown Goal and Unknown Skill, predicts the green goal as the intent. Since in both cases the user's desired trajectory passes near the green goal, PBA erroneously takes control and moves the user towards it, requiring the human to counteract the robot's controls to try to accomplish the task.

In the middle plots for each of the misspecified tasks, we observe for \algabbr after learning, the newly-learned intents receive confidence estimates which increase as the robot is able to observe the user, and thus \algabbr contributes more to the control as it becomes confident.



\begin{figure*}
\centering
\begin{subfigure}{.3\textwidth}
  \centering
  \includegraphics[width=\textwidth,clip=false]{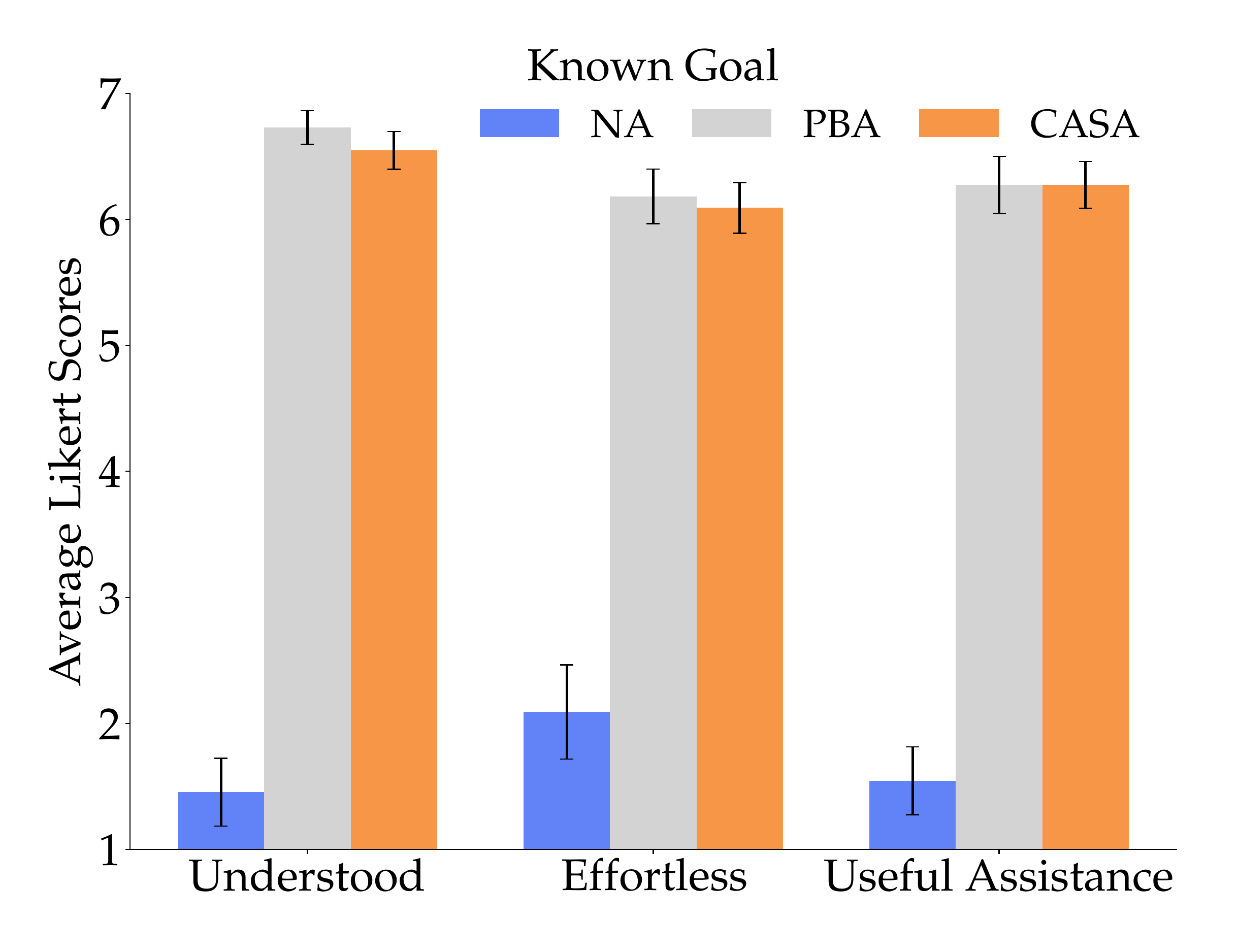}
\end{subfigure}
\centering
\begin{subfigure}{.3\textwidth}
  \centering
  \includegraphics[width=\textwidth,clip=false]{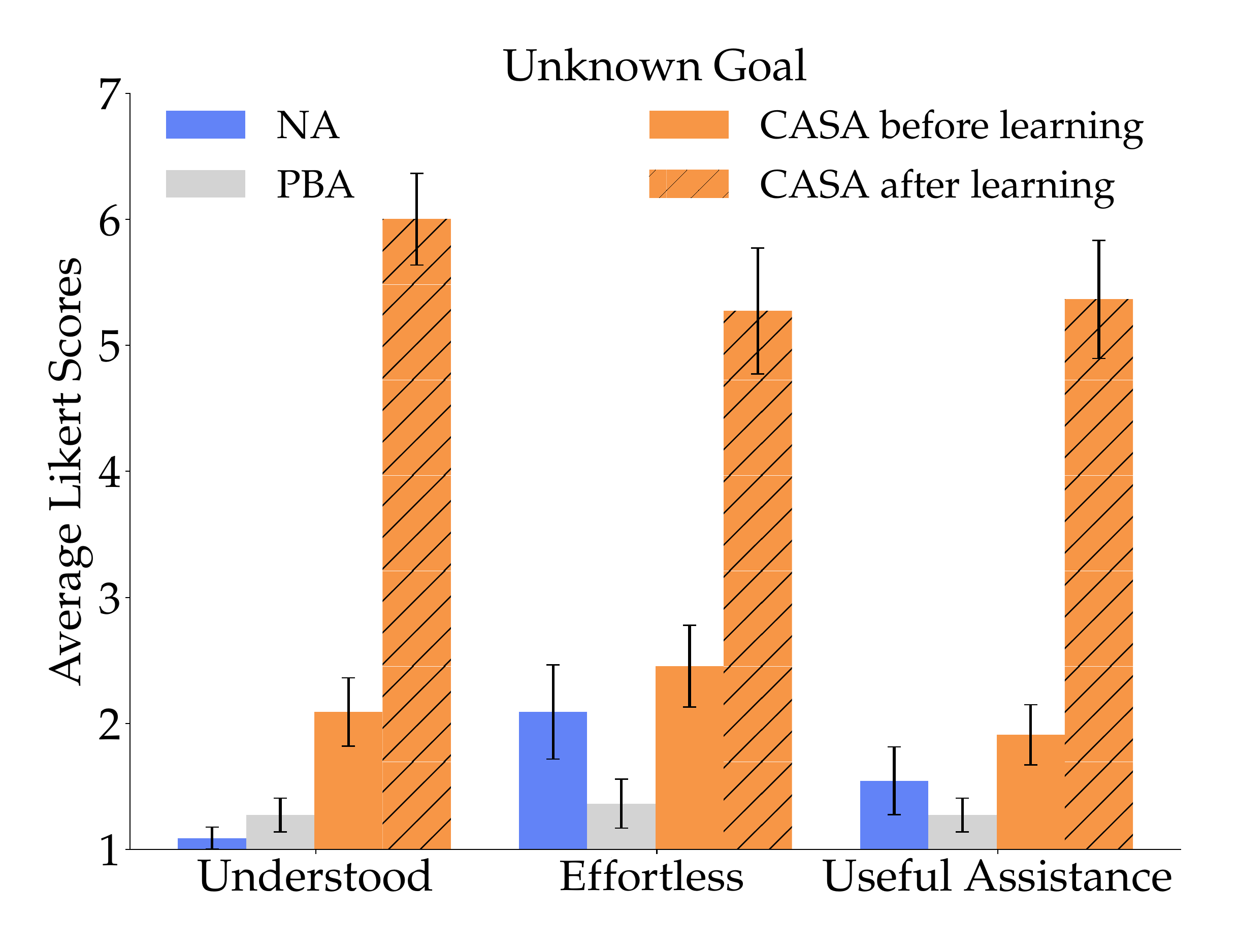}
\end{subfigure}
\centering
\begin{subfigure}{.3\textwidth}
  \centering
  \includegraphics[width=\textwidth]{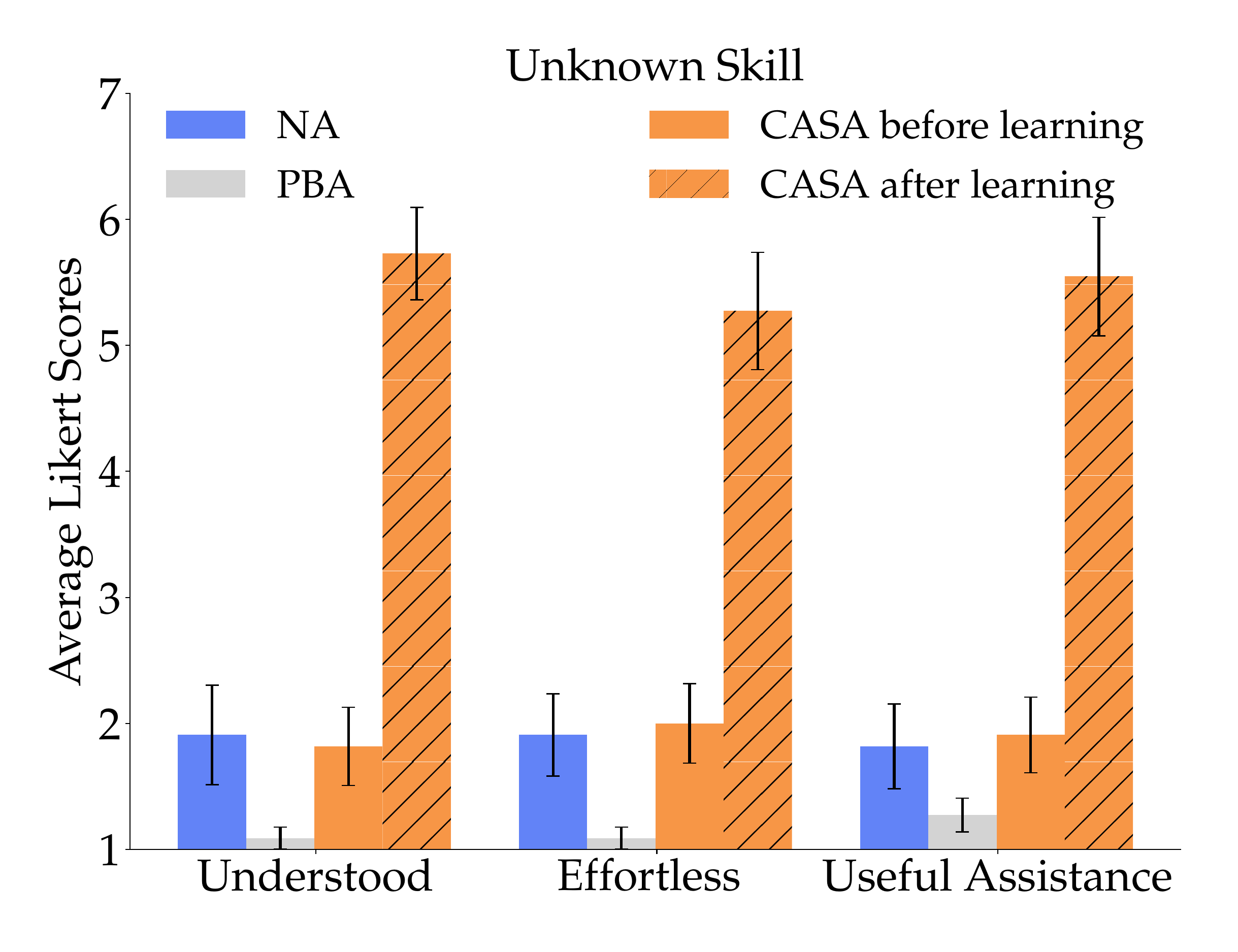}
\end{subfigure}
\caption{Subjective user study results. When there is no misspecification (left), our method is not inferior to PBA, whereas when there is misspecification (center, right), the participants prefer our method after learning a new intent.}
\label{fig:likert}
\vspace*{-0.5cm}
\end{figure*}

%% file: 4_expertstudy.tex
\section{User Study}
\label{sec:userstudy}


We now present the results of our user study, testing how well our method can assist non-expert users.

\subsection{Experimental Design}

Due to the COVID-19 pandemic, we were unable to perform an in-person user study with a physical robot. Instead, as described in Sec. \ref{sec:expert}, we replicated our lab set-up in a pybullet simulator~\cite{coumans2019} in which users can teleoperate a 7 DoF JACO robotic arm using keyboard inputs (Fig. \ref{fig:case_study}). 

We split the study into four phases: (1) familiarization, (2) no misspecification, (3) misspecification before learning, and (4) misspecification after learning. First, we introduce the user to the simulation interface by asking them to perform a familiarization task. In the next phase, we tested the Known Goal task. In the third phase, we tested the two misspecified tasks, Unknown Goal and Unknown Skill, then asked participants to provide 5 demonstrations for each intent. Finally, in the fourth phase, we retested the misspecified tasks using cost functions learned from the demonstrations.

\textbf{Independent Variables:} For each experiment, we manipulate the \textit{assistance method} with three levels: no assistance (NA), policy blending assistance (PBA) \cite{dragan2013blending}, and \algname (\algabbr). For Unknown Goal and Unknown Skill, we compared our method before and after learning new intents against the NA and PBA baselines. 

\textbf{Dependent Measures:} 
Before each task, we displayed an exemplary reference trajectory to help participants understand their objective.
As such, for our objective metrics, we measured 
\textit{Error} as the sum of squared differences between the intended and executed trajectories,
\textit{Efficiency Cost} as the sum of squared velocities across the executed trajectory,
and \textit{Effort} as the number of keys pressed.
To assess the users' interaction experience, we administered a subjective 7-point Likert scale survey, asking the participants three questions: (1) if they felt the robot understood how they wanted the task done, (2) if the robot made the interaction more effortless, and (3) if the assistance provided was useful.

\noindent\textbf{Participants:}
We used a within-subjects design and counterbalanced the order of the assistance methods.
We recruited 11 users (10 male, aged 20-30) from the campus community, most of whom had technical background.

\noindent\textbf{Hypotheses:}

\noindent\textbf{H1:} If there is no misspecification, assisting with \algabbr is not inferior to assisting with PBA, and is superior to NA.

\noindent\textbf{H2:} If there is misspecification, assisting with \algabbr before learning is more accurate, efficient, and effortless than with PBA and not inferior to NA.

\noindent\textbf{H3:} If there is misspecification, assisting with \algabbr after learning is more accurate, efficient, and effortless than NA.

\noindent\textbf{H4:} If there is misspecification, participants will believe the robot understood what they want, feel less interaction effort, and find the assistance more useful with \algabbr after learning than with any other baseline.

\subsection{Analysis}

\noindent\textbf{Objective.}
Fig. \ref{fig:study_objective} summarizes our main findings. For Known Goal, which is well-specified, CASA does no worse than PBA and better that NA in terms of relative effort and error. 
We confirmed this by running an ANOVA, finding a significant main effect for the method ($F(2,30) = 104.93, p < .0001$ for effort; $F(2,30) = 8.93, p = .0009$ for error). In post-hoc testing, a Tukey HSD test revealed that CASA is significantly better than NA ($p < .0001$ for effort, $p = .0013$ for error). 
We also performed a non-inferiority test~\cite{lesaffre2008noninferiority}, and obtained that CASA is non-inferior to PBA within a margin of $0.065$ for effort, $0.025$ for efficiency, and $0.26$ for error. 
These findings are in line with H1 and were expected, since the robot should have no problem handling known intents.

For the two misspecified tasks, we first ran an ANOVA with the method (CASA before learning, NA, and PBA) as a factor, and the task as a covariate, and found a significant main effect ($F(2,62)=11.8255, p<.0001$ for effort; $F(2,62)=6.119, p=.0038$ for error). A Tukey HSD revealed that CASA is significantly better than PBA ($p=.0005$ for effort, $p=.005$ for error). We also ran a non-inferiority test, and obtained that CASA is non-inferior to NA within a margin of $0.035$ for effort, $0.02$ for efficiency, and $1.4$ for error for Unknown Goal, and $0.03$ for effort, $0.09$ for efficiency, and $4.5$ for error for Unknown Skill.
For both unknown tasks, CASA before learning is essentially indistinguishable from NA since a low $\hat\beta_{\goal^*}$ would make the robot rely on direct teleoperation.
Both the figure and our statistical tests confirm H2, which speaks for the consequences of confidently assisting for the wrong intent.

For efficiency cost, we did not find an effect, possibly because Fig.~\ref{fig:study_objective} shows that PBA is more efficient for the Unknown Skill task than other methods. 
Anecdotally, PBA forced users to an incorrect goal thus preventing them from pouring, which explains the lower efficiency cost.
By having a high arbitration for the wrong intent, PBA can cause a smooth trajectory, since it lowers the control authority of the possibly-noisy human inputs. However, this trajectory does not accomplish the task. When running an ANOVA for each of the tasks separately, we found a significant main effect for the method for Unknown Goal ($F(2,30)=9.66, p=.0006$), and a post-hoc Tukey HSD revealed CASA is significantly better than PBA ($p=.0032$), further confirming H2.

Lastly, we looked at the performance with CASA after learning the new intents.
For Unknown Goal, a simple task, the figure shows that CASA after learning doesn't improve efficiency and error, but it does reduce relative effort when compared to NA. 
For Unknown Skill, a more complex task, CASA after learning outperforms NA. This is confirmed by an ANOVA with the method (NA, CASA after learning) as the factor, where we found a significant main effect ($F(1,41)=53.60, p<.0001$ for effort; $F(1,641)=8.6184, p=.0054$ for efficiency cost), supporting H3.

\noindent\textbf{Subjective.}
We show the average Likert survey scores for each task in Fig. \ref{fig:likert}. In line with H1, for the Known Goal task, users thought the robot under both PBA and \algabbr had a good understanding of how they wanted the task to be done, made the interaction more effortless, and provided useful assistance. The results are in stark contrast to NA, which scores low on all those metrics. For Unknown Goal and Unknown Skill, all methods fare poorly on all questions except for CASA after learning, supporting H4.

%% file: 5_discussion.tex
\section{Conclusion}
\label{sec:discussion}

In this paper, we formalized a confidence-aware shared autonomy process where the robot can adjust its assistance based on how confident it is in its prediction of the human intent. We introduced an approximate solution for estimating this confidence, and demonstrated its effectiveness in adjusting arbitration when the robot's intent set is misspecified and enabling continual learning of new intents.

While our confidence estimates tolerated some degree of suboptimal user control, an extremely noisy operator attempting a known intent might instead appear to be performing a novel intent. Moreover, due to COVID, we ran our experiments in a simulator, which does not replicate the difficulty inherent in teleoperating a real manipulator via a joystick interface.
Despite these limitations, we are encouraged to see robots have a more principled and robust way to arbitrate shared autonomy, as well as decide when they need to learn more to be better teammates.
We look forward to applications of our confidence-based ideas beyond manipulation robots, to semi-autonomous vehicles, quadcopter control, or any other shared autonomy scenarios.